\documentclass[10pt,twocolumn,letterpaper]{article}

\usepackage{cvpr}              %

\usepackage{lipsum}
\usepackage{cuted}
\usepackage{comment}
\usepackage{multirow}
\usepackage{tikz}
\usepackage[table]{xcolor}
\definecolor{mygray}{gray}{0.85}

\definecolor{lowpower}{rgb}{1, 0.99, 0.94}
\definecolor{lbpower}{rgb}{1, 0.95, 0.9}
\definecolor{hbpower}{rgb}{1, 0.9, 0.85}
\definecolor{highpower}{rgb}{1, 0.85, 0.8}

\usepackage{threeparttable}
\usepackage{colortbl}
\usepackage{wrapfig}
\usepackage{comment}

\usepackage{fixmetodonotes}

\usepackage[dvipsnames]{xcolor}
\usepackage{pifont}%
\newcommand{\cmark}{\ding{51}}%
\newcommand{\xmark}{\ding{55}}%

\newcommand{\m}[2]{\textit{#1}$^{[#2]}$}
\newcommand{\mm}[1]{\textit{#1}}
\newcommand{\map}[0]{mAP$^{50}$}
\newcommand{\method}{LiDAS\xspace}

\newcommand{\andreic}[1]{}

\newcommand{\simonc}[1]{}

\newcommand{\amandinec}[1]{}

\definecolor{cvprblue}{rgb}{0.21,0.49,0.74}
\usepackage[pagebackref,breaklinks,colorlinks,allcolors=cvprblue]{hyperref}

\usepackage[accsupp]{axessibility}  %

\def\paperID{} %
\def\confName{CVPR}
\def\confYear{2026}

\title{LiDAS: Lighting-driven Dynamic Active Sensing for Nighttime Perception} %

\author{\hspace{-30pt}Simon de Moreau\\
\hspace{-30pt}Valeo - Mines Paris PSL\\
{\hspace{-30pt}\tt\scriptsize simon.de\_moreau@minesparis.psl.eu}
\and
\hspace{-20pt}Andrei Bursuc\\
\hspace{-20pt}valeo.ai\\
{\hspace{-20pt}\tt\scriptsize andrei.bursuc@valeo.com}
\and
\hspace{-20pt}Hafid El Idrissi\\
\hspace{-20pt}Valeo\\
{\hspace{-20pt}\tt\scriptsize hafid.el-idrissi@valeo.com}
\and
\hspace{-20pt}Fabien Moutarde\hspace{-30pt}\\
\hspace{-20pt}Mines Paris PSL\hspace{-30pt}\\
{\hspace{-20pt}\tt\scriptsize fabien.moutarde@minesparis.psl.eu\hspace{-30pt}}
}

\begin{document}
\maketitle

\begin{strip}

    \centering
    
    \vspace{-50pt}
    \includegraphics[width=0.98\textwidth]{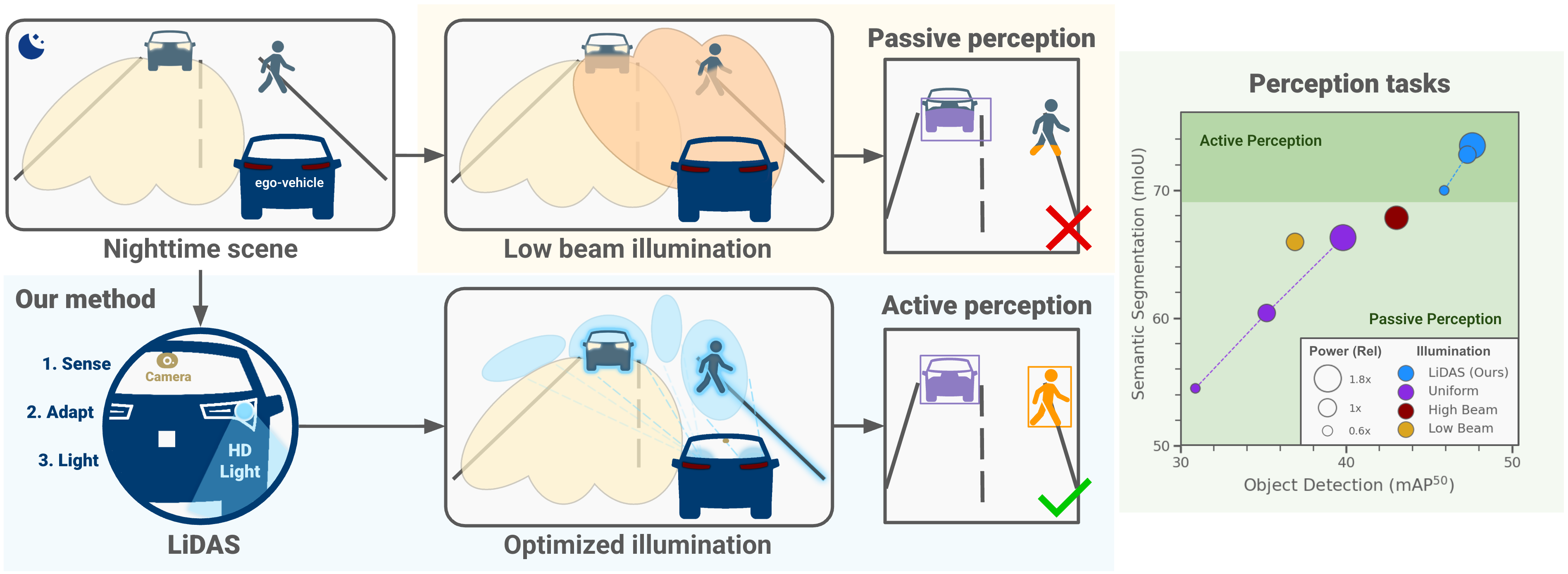}
    \captionof{figure}{
    \textbf{Lighting‑driven Dynamic Active Sensing (LiDAS)} is a bolt‑on active‑illumination system that converts high‑definition headlights into a vision actuator, projecting light where it most aids perception networks.  It improves frozen, daytime‑trained object detection and semantic segmentation models while reducing power consumption.
    }
    \label{fig:concept-fig}
    \vspace{-3pt}
\end{strip}

\begin{abstract}
Nighttime environments pose significant challenges for camera-based perception, as existing methods passively rely on the scene lighting.
We introduce \underline{Li}ghting-driven \underline{D}ynamic \underline{A}ctive \underline{S}ensing (LiDAS), a closed-loop active illumination system that combines off-the-shelf visual perception models with high‑definition headlights. Rather than uniformly brightening the scene, LiDAS dynamically predicts an optimal illumination field that maximizes downstream perception performance, i.e., decreasing light on empty areas to reallocate it on object regions.
LiDAS enables zero-shot nighttime generalization of daytime-trained models through adaptive illumination control. %
Trained on synthetic data and deployed zero‑shot in real‑world closed‑loop driving scenarios, LiDAS enables +18.7\% mAP$^{50}$ and +5.0\% mIoU over standard low‑beam at equal power. It maintains performances while reducing energy use by 40\%. LiDAS complements domain‑generalization methods, further strengthening robustness without retraining. By turning readily available headlights into active vision actuators, LiDAS offers a cost‑effective solution to robust nighttime perception. Project page: \url{https://simondemoreau.github.io/LiDAS/}

\end{abstract}

\section{Introduction}
\label{sec:intro}

Reduced visibility at night contributes to a significant proportion of severe crashes, despite lower traffic volumes \cite{plainis2006road,IIHS}. 
Therefore, ensuring robust performance under low-light conditions is a critical safety requirement for camera-based autonomous systems.
While cameras are cost‑effective and widely deployed, their nighttime performance degrades on unlit and weakly lit roads \cite{wei2025reinpp}. Standard pipelines rely passively on the available scene lighting, but this is a design choice, not a necessity. 
Modern vehicles are able to control where they illuminate and at which intensity.

Vision models trained on daytime data often fail in low-light conditions. While domain adaptation and generalization \cite{hoyer2023mic,kay2025align,li2025clip,yun2025soma} can mitigate this issue, it remains ineffective when scenes are severely underexposed or shift significantly from the training distribution. Alternative sensors (\eg, LiDAR, radar, thermal camera) alleviate some failure modes \cite{ando2023rangevit,zhao2024unimix,huang2025l4dr,bhadoriya2022vehicle,altay2022use} but increase cost and are rarely available in low to mid-cost vehicles. Prior attempts to leverage headlights for perception remain task‑specific or limited to theoretical studies with no real-world evaluation, reducing their generality and practical impact \cite{demoreau2024led,waldner_energy-efficient_2022}.
\newpage
We introduce \underline{Li}ghting-driven \underline{D}ynamic \underline{A}ctive \underline{S}ensing (LiDAS)
that couples standard vision models with high‑definition (HD) headlights enabling real-time, perception-driven control of the lighting. 
Rather than projecting a fixed beam, LiDAS dynamically allocates illumination to scene regions that most benefit downstream tasks such as object detection and semantic segmentation. The method operates in a \emph{closed loop} manner: perceive, adapt the illumination field, project it onto the scene, then perceive again. 
LiDAS aims to 
make the next frame more informative for the existing perception stack. It brings light where it is most needed without increasing 
energy use.

Our work focuses on nighttime driving in weakly lit roads where most of crashes are reported. \method works with frozen downstream perception models, so performance gains come solely from illumination control. Consequently, our method benefits from models trained on large-scale daytime datasets and avoids overfitting to limited nighttime data. In photorealistic simulation and on a closed-loop evaluation using a real vehicle‑mounted prototype, \method improves perception performances while matching or reducing power relative to conventional low-beam (LB) and high-beam (HB). When ambient lighting suffices, it preserves accuracy while saving energy, improving autonomy. Our illumination model informs multiple tasks simultaneously, highlighting the ability of our active lighting to enhance the entire perception stack.
Unlike heuristic adaptive driving beams that react to hand‑crafted saliency or simple object cues \cite{de_charette_fast_2012,fleet_programmable_2014}, LiDAS directly targets downstream task loss. Compared to active sensing approaches that requires additional sensors \cite{ando2023rangevit,zhao2024unimix,huang2025l4dr}, \method leverages cameras and HD headlights already available on modern cars, making the approach practical and cost‑effective.

\noindent Our contributions can be summarized as follows:
\begin{itemize}
    \item We propose \method, a real‑time, power‑efficient active illumination system that closes the perception-illumination loop and allocates light to maximize zero‑shot perception, without task‑specific retraining.  
    \item In simulation, our method improves detection ($+10.4$\% \map) and semantic segmentation ($+6.8$\% mIoU). It matches LB performance with $\!-\!40$\% power.
    \item Our method transfers zero‑shot to a real‑time, car‑mounted prototype and, in closed loop, enhances detection by $+18.7$\% \map and segmentation by $+5.0$\% mIoU, demonstrating real‑world applicability.
    \item \method improves frozen downstream models unseen during training and is compatible with domain‑generalization methods, supporting safer nighttime perception.
\end{itemize}

By focusing available illumination power where it matters most, \method turns headlights into an actuator for vision, improving camera-only perception in the dark without additional sensors or modifications to existing models. 

\section{Related Work}
\label{sec:rw}

\paragraph{Domain adaptation and generalization.}
Domain adaptation (DA) and domain generalization (DG) seek to maintain accuracy under distribution shift. Recent DA methods deliver strong sim‑to‑real and day-to-night improvements \cite{zhou2022multi, hoyer2022daformer,hoyer2022hrda,hoyer2023mic,yang2024micdrop,kay2025align,wei2023disentangle}. DG enhances robustness to distribution shifts and adverse weather through augmentation, normalization, feature‑space alignment \cite{hsu2020progressive,li2025clip,vidit2023clip,lee2024object,wu2024g,danish2024improving}, and increasingly relies on parameter‑efficient fine‑tuning to adapt vision foundation models while preserving general knowledge \cite{hu2022lora,liu2024dora,wei2024stronger,wei2025reinpp,yun2025soma}. In particular, SoMA \cite{yun2025soma} tunes only the minor singular components from a singular value decomposition of pre‑trained weights, achieving strong generalization in both segmentation and detection. While these approaches 
boost performance on nighttime datasets, they %
cannot overcome inherent sensor limits and still fail
in severely under‑exposed scenes. LiDAS complements DA/DG by actively controlling illumination to shift observations toward the training distribution, thereby boosting frozen downstream models.

\vspace{-5pt}

\paragraph{Specialized sensors.}
Standard cameras require sufficient ambient light, whereas active sensors such as LiDAR \cite{ando2023rangevit,zhao2024unimix,li2023pillarnext} and radar \cite{huang2025l4dr,paek2022k,liu2023smurf} are insensitive to illumination, 
a valuable feature for nighttime perception. 
Infrared cameras are also used for nighttime perception \cite{munir2022exploring,bhadoriya2022vehicle,altay2022use} due to their sensitivity to heat signatures. %
Other works \cite{huang2025l4dr,wang2023bi,yu2023benchmarking} fuse multiple sensors to improve robustness. However, all these approaches require additional specialized hardware, increasing system cost and limiting deployment in mainstream vehicles. 
In contrast, our method exploits only the existing camera and leverages the HD headlights already present in most modern vehicles. This offers a practical and low cost solution for improved nighttime safety.

\vspace{-5pt}

\paragraph{Lighting for perception.}
Early works demonstrated that high‑definition headlights can mitigate glare while preserving driver perception, even under adverse conditions \cite{de_charette_fast_2012, fleet_programmable_2014}. More recently, hardware‑in‑the‑loop simulation frameworks have been introduced to prototype and evaluate novel HD headlight functions \cite{waldner_hardware---loop-simulation_2019, waldner_digitization_2020, waldner_optimal_2021}. Building on this line, \cite{waldner_energy-efficient_2022} provided a theoretical analysis of the minimum illumination required for object detection to reduce energy consumption, though their study is restricted to a single simulated scene. 
\cite{demoreau2024led} showed that imposing a fixed HD illumination pattern 
improves depth estimation, but it does not exploit the dynamic, scene‑adaptive capabilities of modern headlights. In contrast, LiDAS actively adapts the illumination field to the scene to improve perception performance, remarkably while reducing energy consumption.

\begin{figure*}

    \centering
    \includegraphics[width=0.9\textwidth]{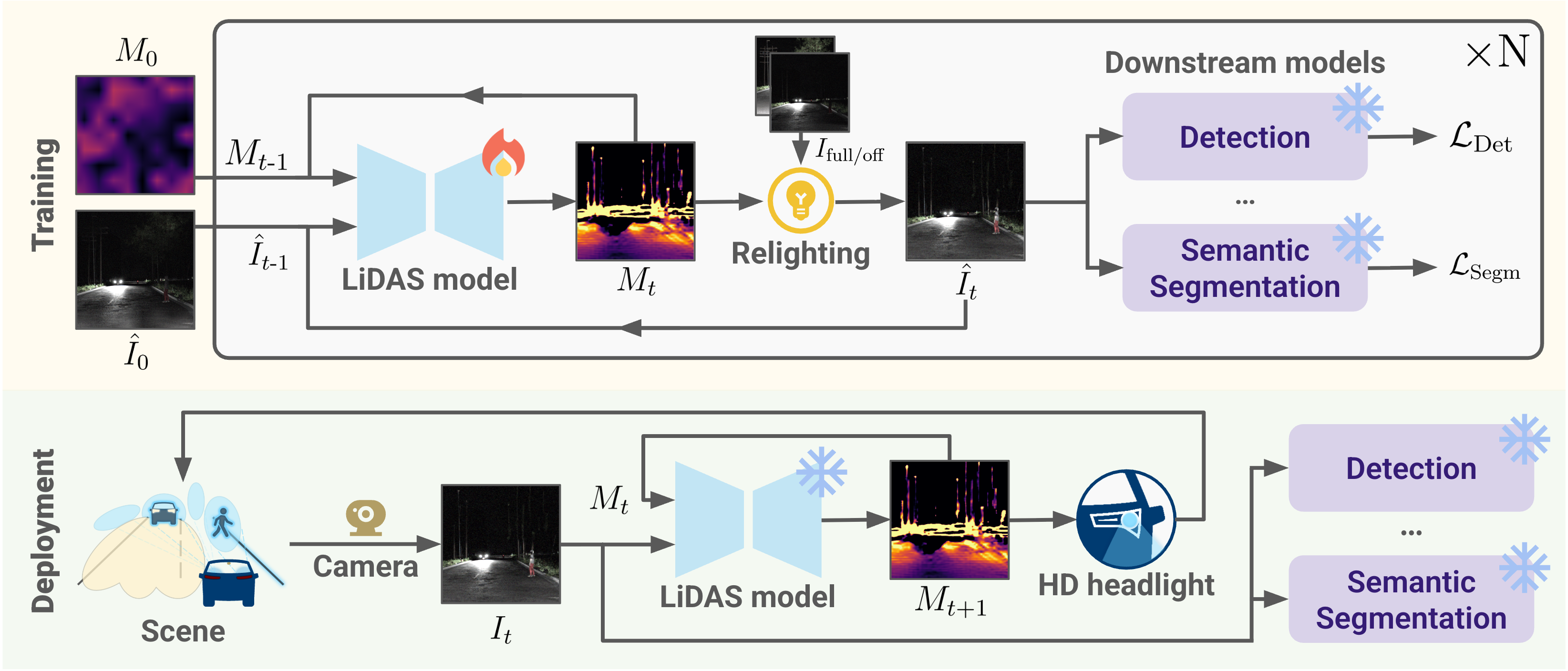}
    \caption{
    \textbf{Training and deployment.} \method learns an illumination policy for active night vision. During training (top), the model predicts a light field $M_t$, which our differentiable relighting operator uses to synthesize the relit image $\hat{I}_t$. Frozen, daytime‑trained downstream heads provide task losses that supervise the policy. We unroll several refinement steps to mirror the closed‑loop setting. At deployment (bottom), the camera observes the lit scene, \method adapts the light field $M_t$, and the HD headlight projects it back onto the scene, improving perception on the next camera frame. At runtime, any models can serve as the downstream heads.
    }
    \label{fig:method-overview}
\end{figure*}

\section{Method}
\label{sec:method}

We learn a vision-driven illumination policy that produces a relevant light field for the current scene. %
Our goal is to improve downstream perception by actively allocating light to the most informative regions of the scene. At training time, our \method model is trained end-to-end through a differentiable relighting operator. At deployment, it drives a real HD headlight, which acts similarly to a video projector, and the camera observes the physical effect. An overview is shown in \cref{fig:method-overview}.

\subsection{Task-driven guidance}
\label{sec:task_guidance}
We adopt an end-to-end, active-sensing approach in which our illumination model is guided by task feedback from frozen perception heads. The relit scene is evaluated using three COCO-pretrained \cite{lin2015microsoft} detectors (YOLO11L, YOLOv8L, YOLOv8L-Worldv2 \cite{Ultralytics_2023}) and a Cityscapes-pretrained \cite{cordts2016cityscapes} segmenter (Mask2Former \cite{cheng2021mask2former}). The training loss is computed as the weighted sum of their task losses.
We found out that, a patch-oriented task like detection promotes localized, contrast-boosting light allocations, while scene-level semantics favors spatially broad light and fine additions on salient regions. In \cref{sec:multitask}, we show that combining multiple tasks and heads regularizes the guidance.

\subsection{Differentiable relighting operator}
\label{sec:relighting}
To perform our end-to-end training, we require a fast, differentiable surrogate model that captures how projected light transforms into the camera image. Our relighting operator linearly interpolates between two renders:
a fully illuminated image $I_{\text{full}}$ (headlights at max power across the field), and
a no-headlight image $I_{\text{off}}$.
Given an image-space light field $M\!\in\![0,1]^{H\times W}$, the relit image $\hat{I}$ is
\begin{equation}
\hat{I_t} \;=\; I_{\text{full}} \odot M_t \;+\; I_{\text{off}} \odot \big(1 - M_t\big),
\label{eq:relight}
\end{equation}

\noindent where $\odot$ denotes elementwise multiplication. \cref{eq:relight} preserves gradients \wrt $M_t$ and simplify the light projection,
 enabling efficient end-to-end training.

For classic LB and HB baselines, we model both scene geometry and headlamp photometry. Using per‑pixel depth, we lift image points to 3D and transform them into the headlight frame via known extrinsics. Treating the headlight as a pinhole projector, we project 3D points to pixels using its intrinsics. We then sample the measured angular intensity distribution $\Phi_{\text{LB/HB}}$ from a real high‑end vehicle and apply the resulting illumination back into the image.

The two-image interpolation naturally includes ambient and non-ego light (present in both $I_{\text{off}}$ and $I_{\text{full}}$), and allows exhaustive exploration of relighting configurations from a single pair at negligible cost. Limitations remain: material reflectivity are not evaluated and saturated regions in $I_{\text{full}}$ may not recover detail when decreasing $M$. Nevertheless, it provides the correct signal indicating where light helps or harms downstream task performance, which is sufficient to train our model. Limitations of the operator are discussed in the suppl. material.

\subsection{Training strategy}
\label{sec:training}

\paragraph{Initialization.}
\label{sec:init}
We start from a randomized illumination $M_{0}$ to encourage lighting-agnostic behavior. We draw blockwise-constant noise with block sizes in $[20,80]$ pixels, intensity is scaled to match the energy budget. With probability $p{=}0.5$, we use a fully black field $M_{0}\!=\!0$ so the model learns to re-illuminate previously dark regions when necessary. The starting image $\hat{I}_0$ is obtained using \cref{eq:relight}.

\paragraph{LiDAS model.}
\label{sec:lidas_model}
The model's inputs are an RGB image $I\!\in\!\mathbb{R}^{H\times W\times 3}$, the previous illumination map $M_{t-1}\!\in\![0,1]^{H\times W}$, and a CoordConv \cite{liu2018intriguing} inspired 
channel
$C=(x,y)\!\in\![0,1]^2$. Conditioning on $M_{t-1}$ helps the network distinguish self-induced illumination from ambient light. %
CoordConv provides coarse spatial context that guides the light distribution (see \cref{sec:ablations}).

The architecture is an encoder-decoder with skip connections: four downsampling stages to $1/16$ resolution, followed by two upsampling stages to $1/4$, and a final resize to full resolution. Our network learns residual corrections of the light field rather than absolute illumination.
The prediction head outputs a residual update $\Delta M_t \in [-1,1]^{H\times W}$, which we add to the previous map $M_{t-1}$ and project element-wise onto the valid range:
\begin{equation}
\widetilde{M}_t \;=\; 
\min\!\big(\max(M_{t-1} + \Delta M_t,\, 0),\, 1\big).
\label{eq:residual}
\end{equation}

\paragraph{Scheduling the energy budget.}
\label{sec:energy_schedule}
We enforce a per-frame power budget by normalizing the field $\widetilde{M}_t$ to a target $\eta$:
\begin{equation}
M_t \;=\; \frac{\eta}{\max(\epsilon,\;\bar{m}_t)}\, \widetilde{M}_t,
\label{eq:budget}
\end{equation}
with $\bar{m}_t$ the mean intensity of $\widetilde{M}_t$ and a small $\epsilon\,{>}\,0$ for numerical stability. To encourage early prioritization and later refinement, we linearly ramp the budget across epochs:
\begin{equation}
\eta(e) \;=\; \eta_{\text{final}}\!\left(\,\alpha \;+\; (1-\alpha)\,\frac{e}{E_{\max}}\,\right),
\label{eq:ramp}
\end{equation}
with $E_{\max}$ the number of training epochs, $e$ the current epoch and $\alpha=10\%$. %
The resulting $M_t$ is used by the relighting operator during training and by the HD headlight at runtime.

\paragraph{Sequential refinement.}
\label{sec:autoregressive}
Although supervision is performed on single images, deployment is closed-loop: LiDAS model acts, then observes its own illumination in the next frame. To match this setting, we unroll the model illumination on the same training image for $N\!=\!40$ iterations. For $t\!=\!0,\dots,N\!-\!1$ we apply the relighting operator \cref{eq:relight} and feed the result back to the model:

\begin{equation}
M_{t+1} \;=\; \text{LiDAS} \big(\hat{I}_{t},\, M_{t},\, C\big).
\label{eq:policy_update}
\end{equation}
Downstream task losses are backpropagated $K\!=\!5$ times within the first iteration and 4 randomly sampled steps. Gradients do not propagate across iterations: each step treats $(\hat{I}_{t}, M_t)$ as constants for the next, which encourages stability under the model’s own illumination while avoiding full backpropagation through time. 
This strategy promotes steady performance over long horizons in closed loop (see suppl. material).

\paragraph{Runtime deployment.} 
At deployment, only the LiDAS model is required. It runs as a bolt‑on module that interfaces with the vehicle’s existing headlights and camera, leaving the native perception stack unchanged. Because LiDAS operates in the \emph{image} frame, the predicted light field is mapped to headlight pixels via a precomputed warp obtained from a one-time camera-headlight calibration.

\section{Experimental Setup}
\label{sec:exp_setup}
\subsection{Datasets}
\paragraph{Synthetic.}
We use the Applied Intuition photorealistic simulator \cite{Applied} to render nighttime scenes with street lighting disabled, replicating the worst-illumination conditions, where most crashes occur. We include pedestrians, bicyclist, motorcyclist and cars. Vehicles have their headlights on with real measured LB/HB photometry. %
Four maps are used for training and validation and two separated maps for testing. Maps are evenly split between city and roads. Our dataset includes 2,250/250/1,000 images for train/val/test. Unless specified, the weather is clear with ambient light ranging from dark to full‑moon.

\begin{figure*}

    \centering
    \includegraphics[width=1\textwidth]{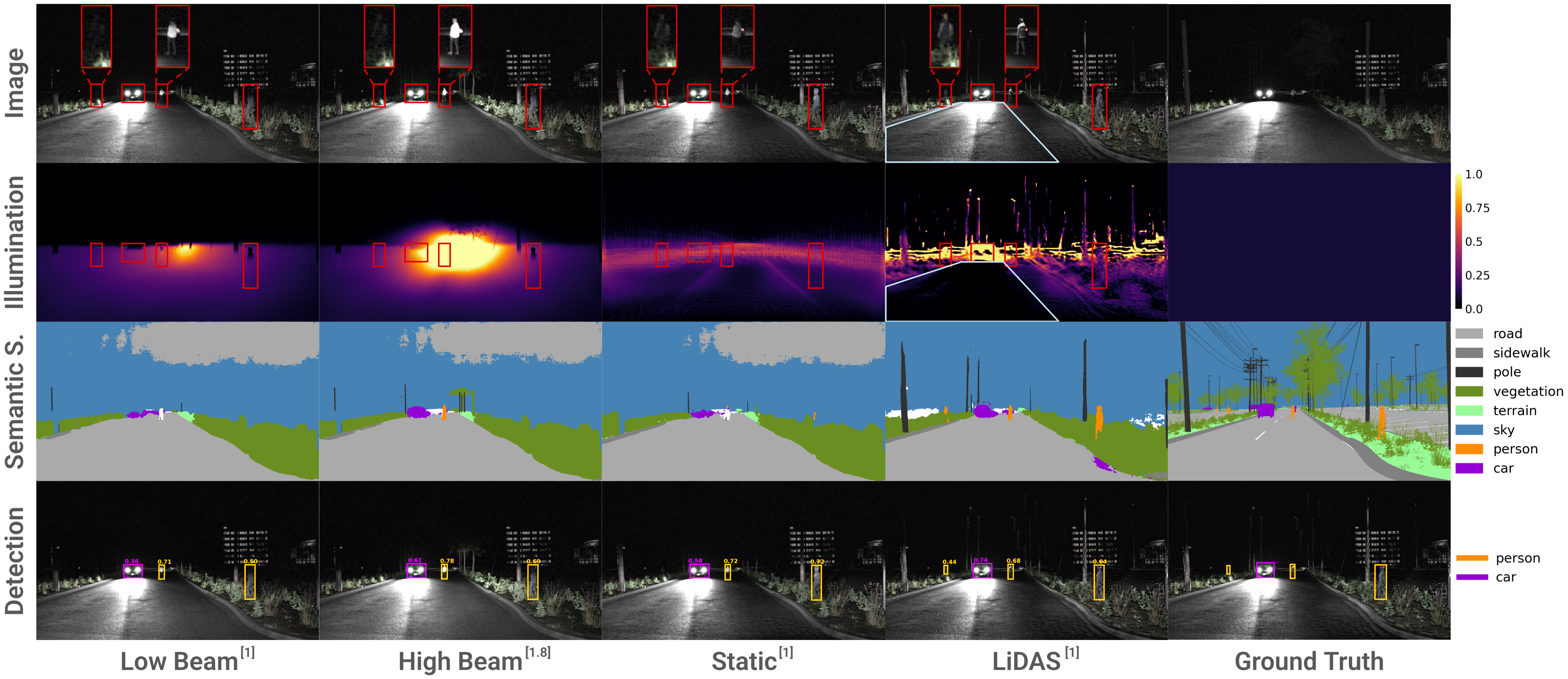}
    \caption{
    \textbf{Qualitative results.} We show YOLO11L (detection) and Mask2Former (semantic segmentation) predictions under different illumination policies. Only \m{LiDAS}{1} detects the left‑hand pedestrian and its segmentation map has well‑defined objects and no sky artifacts. \m{LiDAS}{1} lights objects of interest while leveraging ambient illumination, reducing power over the other car’s headlights (blue 
    polygon) and the pedestrian’s white coat. \mm{Static} shows that \method typically reduces near-field light avoiding self-glare and reallocates power toward long‑range targets. Method$^{[x]}$ denotes power relative to LB.
    }
    \label{fig:qualitative_main}
\end{figure*}

\paragraph{Real-world.}
We perform zero‑shot on‑vehicle deployment on a professional test tracks. They are populated with 12 certified human targets and 10 vehicles to create oncoming, following, and parked‑vehicle scenarios. All road assets are kept static to ensure run‑to‑run reproducibility. We evaluate unlit rural roads and urban area, with and without street lighting, all under clear weather. Speeds are $\simeq$30\,km/h in urban sections and $>$50\,km/h on rural roads. The test vehicle has HD headlights (320$\times$80\,px) and a front-facing camera. For each illumination mode (LB, HB, and \method under two energy budgets)
we drive the vehicle over a distance of 12\,km. Ground‑truth labels are obtained using LiDAR: 3D point clouds are manually labeled and projected into the camera viewpoint \cite{mathias2025doc}. In this experiment, we consider only the vehicle and person classes.

\subsection{Downstream tasks and metrics}
Our model is trained and evaluated end-to-end with frozen perception models. 
Unless otherwise specified, for detection we report metrics from a COCO-pretrained \cite{lin2015microsoft} YOLO11L \cite{Ultralytics_2023} metrics :
Precision, Recall, \map, mAP$^{50-90}$. For semantic segmentation, we use Mask2Former \cite{cheng2021mask2former} pretrained on Cityscapes \cite{cordts2016cityscapes}, and report mIoU and mAcc. Reported powers are always \emph{relative to LB} as it is the most commonly used headlight setting \cite{IIHS_HB_usage,AAA_HB_usage}.

\subsection{Implementation details}
We train our model for 60 epochs using the AdamW optimizer \cite{loshchilov2017decoupled} with a learning rate of $10^{-4}$, exponential decay ($\gamma=0.96$), and mixed precision. 
Energy is linearly scheduled from \(\eta_0=0.1\times\eta_{\text{final}}\) to \(\eta_{\text{final}}\). 
Our model has 54M parameters. It is trained for ${\sim}$16 hours on a single H100. The inference time is 6.8\,ms on a RTX 4090, inducing only minimal overhead to the pipeline, details in suppl. material.

\subsection{Baselines}
\label{sec:baselines}
We compare our method against real measured \mm{Low Beam} and \mm{High Beam} from high‑end vehicles, which we assume to be the best‑case passive illumination in standard systems. The \mm{Uniform} baseline distributes the available power evenly to brighten the entire scene, while the \mm{No Ego Light} one isolates the contribution of ego illumination to metrics. We include a \mm{Static} pattern computed as the average of \method predictions on the validation dataset. For brevity, we denote configurations as \m{Method}{\text{Power}}. 

\section{Experiments}
\label{sec:exp}
\subsection{Synthetic environment}
\begin{table}[t]
\centering
\caption{\textbf{Performances on synthetic dataset.} Results are reported for YOLO11L (detection) and Mask2former (semantic segmentation). \method consistently surpasses all baselines. Notably, even \m{LiDAS}{0.6}, which saves $40$\% power, outperforms higher-energy methods. It underscores 
the ability of \method to optimize illumination for perception.}

\resizebox{\linewidth}{!}{
    \begin{tabular}{l|c|cccc|cc}
    \toprule
     \multirow{2}{*}{\textbf{Method}} & \multirow{2}{*}{\textbf{Power}} & \multicolumn{4}{c|}{\textbf{Detection}} & \multicolumn{2}{c}{\textbf{Semantic Segmentation}} \\
    
      &  & \textbf{P $\uparrow$} & \textbf{R $\uparrow$} & \textbf{mAP$^{50}$ $\uparrow$} & \textbf{mAP$^{50-90}$ $\uparrow$} & \textbf{mIoU $\uparrow$} & \textbf{mAcc $\uparrow$} \\
      
    \midrule
    No Ego Light & 0   & 43.0 & 19.7 & 21.5 & 11.6 & 47.6 & 65.4 \\
    \rowcolor{lowpower} Uniform     & 0.6 & 53.0 & 27.2 & 30.9 & 17.9 & 54.5 & 76.7 \\
    \rowcolor{lowpower} Static      & 0.6 & 55.5 & 36.4 & 39.6 & 23.9 & 62.0 & 77.4 \\
    \rowcolor{lowpower} \textbf{LiDAS (Ours)} & 0.6 & \textbf{66.0} & \textbf{39.7} & \textbf{45.9} & \textbf{29.1} & \textbf{70.0} & \textbf{84.1} \\    
    \rowcolor{lbpower} Low Beam    & 1 & 62.5 & 30.6 & 36.9 & 21.5 & 66.0 & 80.5 \\
    \rowcolor{lbpower} Uniform     & 1 & 57.0 & 31.0 & 35.2 & 20.7 & 60.4 & 81.0 \\
    \rowcolor{lbpower} Static      & 1 & 65.7 & 35.8 & 42.6 & 26.3 & 68.7 & 82.7 \\        
    \rowcolor{lbpower} \textbf{LiDAS (Ours)} & 1 & \textbf{66.9} & \textbf{40.6} & \textbf{47.3} & \textbf{30.0} & \textbf{72.8} & \textbf{85.6} \\ 
    \rowcolor{hbpower} High Beam   & 1.8 & \textbf{66.8} & 35.6 & 43.0 & 26.8 & 67.9 & 82.4 \\
    \rowcolor{hbpower} Uniform     & 1.8 & 66.0 & 32.8 & 39.8 & 23.8 & 66.3 & 84.2 \\
    \rowcolor{hbpower} Static      & 1.8 & 62.2 & 38.7 & 44.6 & 28.3 & 71.1 & 85.2 \\        
    \rowcolor{hbpower} \textbf{LiDAS (Ours)} & 1.8 & 65.8 & \textbf{41.3} & \textbf{47.6} & \textbf{30.4} & \textbf{73.5} & \textbf{87.0} \\        
    \rowcolor{highpower} Uniform     & 4  & 60.0 & 39.3 & 44.5 & 28.2 & 71.9 & 86.1 \\
    \bottomrule
    \end{tabular}
}

\vspace{-10pt}
\label{tab:main_sim}
\end{table}

\begin{figure*}

    \centering
    \includegraphics[width=1\textwidth]{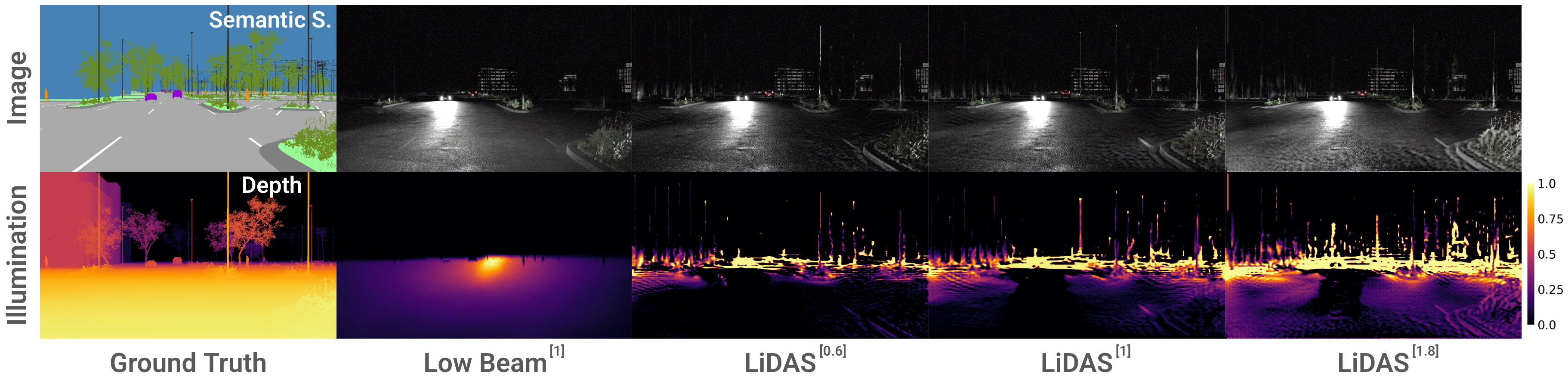}
    \caption{
    \textbf{Budget‑aware illumination.}
    LiDAS learns to prioritize illumination under tight power budgets, focusing only on the most informative regions.
    As the budget grows, it progressively broadens coverage and begins to accentuate finer details that aid global scene understanding. Method$^{[x]}$ denotes power relative to LB.
    }
    \label{fig:qualitative_budget}
    \vspace{-10pt}
\end{figure*}

\paragraph{Quantitative results.}
We report results in \cref{tab:main_sim}.
With the same power as \m{Low Beam}{1}, \m{LiDAS}{1} improves detection by $+10.4$\% \map and semantic segmentation by $+6.8$\% mIoU. It even surpasses \m{High Beam}{1.8} despite the latter using $1.8\times$ more energy. Moreover, with a $40$\% power reduction, \m{LiDAS}{0.6} outperforms all baselines across power levels. Even \m{Uniform}{4}, which consumes the most power and brightens the entire scene, fails to match \method because its lighting suppresses contrast. In supplementary material, we show that \m{LiDAS}{1} achieves leading performance in the 20-60\,m range, a safety‑critical band for autonomous emergency braking. Overall, these results indicate that perception depends less on the total light flux and more on how light is spatially allocated.

\vspace{-5pt}

\paragraph{Qualitative results.}
\cref{fig:qualitative_main} shows that \mm{LiDAS} reduces near‑field illumination, avoiding self‑glare, and reallocates energy toward the horizon where distant objects need it.
\mm{LiDAS} also decreases intensity over already lit regions (\eg, other vehicles’ lights) and sufficiently exposed objects (\eg, pedestrian's white coat), improving local contrast and boundary definition. 
\cref{fig:qualitative_budget} further demonstrates LiDAS’s ability to adapt to its power budget: at low budgets it prioritizes essential, perception‑critical regions, as the budget increases it broadens coverage, and at high budgets it accentuates finer structures that contribute to a holistic understanding of the scene.

\begin{table}[t]
\centering
\caption{\textbf{
Robustness to change of environment.} We assess performance in zero‑shot mode across street‑lit scenes and rain. When roads are correctly lit, all methods perform similarly, showing minimal benefit from ego‑vehicle illumination. Under rain, LiDAS generalizes effectively to the unseen setting and surpasses other approaches.}
\vspace{-5pt}

\resizebox{\linewidth}{!}{
    \begin{tabular}{l|c|cccc|cc}
    \toprule
     \multirow{2}{*}{\textbf{Method}} & \multirow{2}{*}{\textbf{Power}} & \multicolumn{4}{c|}{\textbf{Detection}} & \multicolumn{2}{c}{\textbf{Semantic Segmentation}} \\
      &  & \textbf{P $\uparrow$} & \textbf{R $\uparrow$} & \textbf{mAP$^{50}$ $\uparrow$} & \textbf{mAP$^{50-90}$ $\uparrow$} & \textbf{mIoU $\uparrow$} & \textbf{mAcc $\uparrow$} \\
    \midrule
    \multicolumn{8}{l}{\cellcolor{gray!10}\emph{Urban area
    with street lights}} \\
    No Ego Light    & 0 & 71.4 & 38.0 & 46.5 & 30.1 & 74.0 & 87.0 \\
    \rowcolor{lowpower} \textbf{LiDAS (Ours)} & 0.6 & 71.6 & 39.8 & 47.6 & 30.9 & \underline{80.4} & 88.3 \\ 
    \rowcolor{lbpower} Low Beam    & 1 & \underline{71.7} & 39.8 & \underline{48.3} & 31.0 & 76.6 & 88.2 \\
    \rowcolor{lbpower} \textbf{LiDAS (Ours)} & 1 & 71.6 & 39.9 & 47.4 & \underline{31.2} & \textbf{81.5} & \underline{88.9} \\ 
    \rowcolor{hbpower} High Beam   & 1.8 & \textbf{71.8} & \textbf{41.0} & \textbf{49.1} & \textbf{31.5} & 77.4 & \textbf{89.0} \\
    \rowcolor{hbpower} \textbf{LiDAS (Ours)} & 1.8 & 69.1 & \underline{40.5} & 47.2 & 30.8 & 80.0 & 88.2 \\
        \midrule
    \multicolumn{8}{l}{\cellcolor{gray!10}\emph{Rain}} \\
    \rowcolor{lowpower} \textbf{LiDAS (Ours)} & 0.6 & 52.0 & \textbf{36.8} & 38.2 & 22.2 & 52.9 & 76.4 \\
    \rowcolor{lbpower} Low Beam    & 1 & 53.0 & 26.9 & 29.3 & 17.7 & 57.6 & 81.2 \\
    \rowcolor{lbpower} \textbf{LiDAS (Ours)} & 1 & 55.6 & \underline{36.2} & \underline{39.2} & \underline{23.6} & 59.4 & 81.0 \\ 
    \rowcolor{hbpower} High Beam   & 1.8 & \underline{58.4} & 30.8 & 34.5 & 20.6 & \underline{60.2} & \textbf{83.5} \\
    \rowcolor{hbpower} \textbf{LiDAS (Ours)} & 1.8 & \textbf{63.2} & 35.7 & \textbf{41.5} & \textbf{24.7} & \textbf{65.5} & \underline{82.8} \\   
    \bottomrule  
    \end{tabular}

}

\vspace{-5pt}
\label{tab:env_robustness}
\end{table}

\paragraph{Static vs.\ dynamic illumination.}
\label{sec:static_dynamic}
We construct \m{Static}{x} by averaging \m{LiDAS}{x} predictions, yielding a perception-optimized, yet non-adaptive, light field. As shown in \cref{tab:main_sim}, \m{Static}{1/1.8} outperforms LB/HB but still trails behind the dynamic \m{LiDAS}{1} by $-4.7$\% \map and $-2.9$\% mIoU. Lacking scene adaptivity, \mm{Static} cannot account for road curvature, object location, or reflectivity. Thus, its light allocation is broader and less targeted. This wastes energy that cannot be reallocated to informative regions and leaves some objects under-lit, as the left‑side pedestrian in \cref{fig:qualitative_main}. Overall, this underscores the advantage of active over passive lighting for perception.%

\begin{table}[t]
\centering
\caption{\textbf{Downstream model generalization.} We evaluate YOLOv5 and SegFormer as downstream models unseen during training. LiDAS consistently surpasses baselines, highlighting robust zero‑shot transfer.}
\vspace{-5pt}
\resizebox{\linewidth}{!}{
    \begin{tabular}{l|c|cccc|cc}
    \toprule
     \multirow{2}{*}{\textbf{Method}} & \multirow{2}{*}{\textbf{Power}} & \multicolumn{4}{c|}{\textbf{Detection}} & \multicolumn{2}{c}{\textbf{Semantic Segmentation}} \\
      &  & \textbf{P $\uparrow$} & \textbf{R $\uparrow$} & \textbf{mAP$^{50}$ $\uparrow$} & \textbf{mAP$^{50-90}$ $\uparrow$} & \textbf{mIoU $\uparrow$} & \textbf{mAcc $\uparrow$} \\
    \midrule
    \multicolumn{1}{l|}{\cellcolor{gray!10}\emph{}} & \multicolumn{1}{l|}{\cellcolor{gray!10}\emph{}}& \multicolumn{4}{l|}{\cellcolor{gray!10}\emph{Yolov5L}} & \multicolumn{2}{l}{\cellcolor{gray!10}\emph{Segformer}}\\
    \rowcolor{lowpower} \textbf{LiDAS (Ours)} & 0.6 & \textbf{65.5} & 38.4 & 44.1 & 27.2 & 65.7 & \underline{75} \\ 
    \rowcolor{lbpower} Low Beam    & 1 & 56.7 & 29.6 & 33.8 & 19.8 & 59.9 & 69 \\
    \rowcolor{lbpower} \textbf{LiDAS (Ours)} & 1 & \underline{62.3} & \underline{39.8} & \underline{44.8} & \underline{28.1} & \textbf{66.4} & \textbf{75.6} \\ 
    \rowcolor{hbpower} High Beam   & 1.8 & 56.6 & 37.1 & 40.8 & 25.0 & 62.8 & 72.4 \\
    \rowcolor{hbpower} \textbf{LiDAS (Ours)} & 1.8 & 61.3 & \textbf{41.2} & \textbf{45.7} & \textbf{28.7} & \underline{65.8} & 74.8 \\ 
    \bottomrule
    \end{tabular}
}
\vspace{-15pt}
\label{tab:gen_down}
\end{table}

\subsection{Generalization} %

\paragraph{Environment robustness.}
We evaluate LiDAS in a zero-shot setting under additional environmental conditions by generating 500 extra samples of well-lit urban nighttime scenes and unlit roads with rainy weather. Results in \cref{tab:env_robustness} show that, in correctly illuminated areas, all methods perform similarly, including the \mm{No‑Ego-Light} baseline. It indicates that ego‑vehicle illumination has limited leverage when ambient lighting already dominates and environment-induced glare cannot be avoided. In contrast, on unlit roads, even under rain, \mm{LiDAS} improves detection while maintaining segmentation performance. These results suggest that our model adapts to previously unseen reflective phenomena (\eg, reflections on wet road), selectively allocating light to preserve object salience while avoiding self-glare.

\begin{figure*}

    \centering
    \includegraphics[width=1\textwidth]{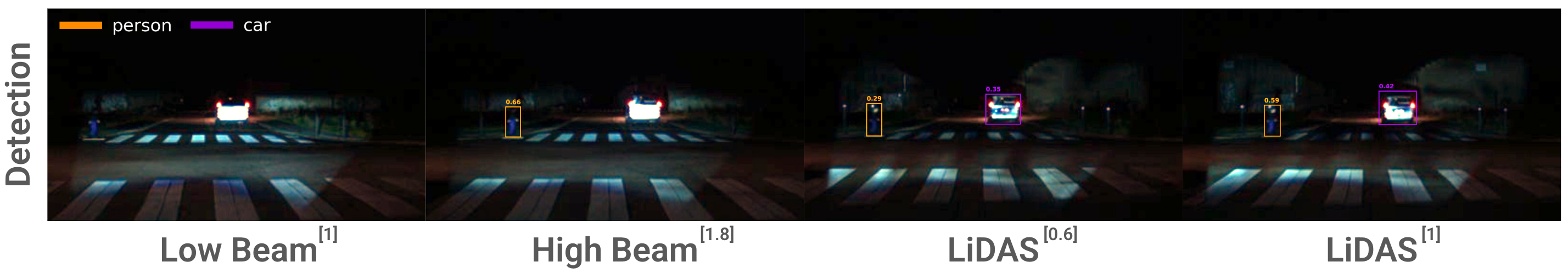}
    \caption{ 
    \textbf{Qualitative results from the closed-loop evaluation on our car-mounted prototype.} We compare YOLOv8L-Worldv2 predictions under classic LB/HB and our LiDAS illumination policy at several energy budgets. LiDAS detects all present objects even at reduced power. It makes the pedestrian fully visible and reduces self‑glare on the white vehicle, improving contrast. Method$^{[x]}$ denotes power relative to LB.
    }
    \label{fig:qualitative_real}
\end{figure*}

\paragraph{Downstream models.}
We assess whether LiDAS, trained jointly with its fixed set of downstream tasks, generalizes to previously unseen models. Without retraining LiDAS, we evaluate YOLOv5L, YOLO12L \cite{Ultralytics_2023}, and SegFormer \cite{xie2021segformer} as downstream heads. As reported in \cref{tab:gen_down} and in the suppl. material, \method consistently improves perception across these models, indicating that the learned illumination policy supports zero‑shot generalization to new models of various size, boosting performance in each case. %

\subsection{Zero-shot sim-to-real transfer}
\label{sec:sim2real}

We perform zero-shot evaluation on real nighttime images using the same frozen downstream models (YOLO \cite{Ultralytics_2023} detectors pretrained on COCO \cite{lin2015microsoft}; Mask2Former \cite{cheng2021mask2former} pretrained on Cityscapes \cite{cordts2016cityscapes}). Both \method and the downstream models are used without any fine-tuning. Additional results on nuImages \cite{caesar2020nuscenes}, BDD100K \cite{yu2020bdd100k} and NightCity \cite{tan2021night} and using DTP \cite{wei2023disentangle} are provided in the suppl. material. %

\paragraph{Closed-loop evaluation.} %
\begin{table}[t]
\centering
\caption{\textbf{Performance in real-world deployment.} Results for YOLOv8L-Worldv2 (detection) and Mask2Former (semantic segmentation). In zero‑shot, closed‑loop operation on a car‑mounted prototype, \method surpasses both baselines and proves deployable in practice, with large improvements on every metric. The \method illumination policy is learned on synthetic data.}
\vspace{-5pt}

\resizebox{\linewidth}{!}{
    \begin{tabular}{l|c|cccc|cc}
    \toprule

     \multirow{2}{*}{\textbf{Method}} & \multirow{2}{*}{\textbf{Power}} & \multicolumn{4}{c|}{\textbf{Detection}} & \multicolumn{2}{c}{\textbf{Semantic Segmentation}} \\
    
      &  & \textbf{P $\uparrow$} & \textbf{R $\uparrow$} & \textbf{mAP$^{50}$ $\uparrow$} & \textbf{mAP$^{50-90}$ $\uparrow$} & \textbf{mIou} $\uparrow$ & \textbf{mAcc} $\uparrow$\\
      
    \midrule
    \multicolumn{8}{l}{\cellcolor{gray!10}\emph{Unlit urban area and rural roads}} \\
    \rowcolor{lowpower} \textbf{LiDAS (Ours)} & 0.6 & \underline{52.5} & \underline{7.9} & \underline{30.2} & \underline{12.0} & 10.7 & 11.0 \\        
    \rowcolor{lbpower} Low Beam & 1 & 30.2 & 1.6 & 15.8 & 4.6 & 10.3 & 10.7  \\
    \rowcolor{lbpower} \textbf{LiDAS (Ours)} & 1 & \textbf{56.9} & \textbf{12.6} & \textbf{34.5} & \textbf{13.2} & \textbf{15.3} & \textbf{15.8}  \\   
    \rowcolor{hbpower} High Beam & 1.8 & 30.6 & 2.4 & 16.0 & 4.6 & \underline{11.1} & \underline{11.5}  \\

    \midrule

    \multicolumn{8}{l}{\cellcolor{gray!10}\emph{Urban area with street lights}} \\
    \rowcolor{lowpower} \textbf{LiDAS (Ours)} & 0.6 & \textbf{54.1} & \underline{21.5} & \underline{37.0} & \underline{15.2} & 22.9 & 23.6  \\        
    \rowcolor{lbpower} Low Beam & 1 & 52.3 & 15.3 & 33.4 & 12.8 & 22.6 & 23.2  \\
    \rowcolor{lbpower} \textbf{LiDAS (Ours)} & 1 & \underline{53.8} & \textbf{24.0} & \textbf{37.9} & \textbf{16.0} & \textbf{23.5} & \textbf{24.5} \\   
    \rowcolor{hbpower} High Beam & 1.8 & 46.4 & 16.5 & 30.6 & 11.2 & \underline{23.0} & \underline{23.8}  \\
    \bottomrule
    \end{tabular}
}

\label{tab:real-world}
\end{table}

We evaluate LiDAS in zero‑shot, real‑time conditions using a car‑mounted prototype, and report results in \cref{tab:real-world} and \cref{fig:qualitative_real}. On unlit roads, \m{LiDAS}{1} yields larger gains than in simulation, with $+18.7$\% \map and $+5.0$\% mIoU over \m{Low Beam}{1}.
Even the energy‑saving \m{LiDAS}{0.6} outperforms \m{High Beam}{1.8}, achieving $+5.5$\% recall. It shows that LiDAS generalizes well in zero‑shot on real data despite being trained solely on synthetic data. It also confirms that our closed‑loop active sensing method effectively improves perception and is sufficiently responsive for operation on a moving vehicle.
Even under street lighting, we observe improvements, likely due to under-lit areas on the track and the real camera’s light sensitivity benefiting from LiDAS’s optimized illumination.

\paragraph{Combination with domain generalization methods.}
\begin{table}[t]
\centering
\caption{\textbf{Combination with domain generalization methods on real-world scenes.} We integrate \method with SoMA. SoMA ensures strong low‑beam performance via domain generalization, and \method delivers additional zero‑shot improvements, confirming the synergy between DG and active‑lighting methods.}
\vspace{-5pt}

\resizebox{\linewidth}{!}{
    \begin{tabular}{l|c|cccc|cc}
    \toprule

     \multirow{2}{*}{\textbf{Method}} & \multirow{2}{*}{\textbf{Power}} & \multicolumn{4}{c|}{\textbf{Detection}} & \multicolumn{2}{c}{\textbf{Semantic Segmentation}} \\
    
      &  & \textbf{P $\uparrow$} & \textbf{R $\uparrow$} & \textbf{mAP$^{50}$ $\uparrow$} & \textbf{mAP$^{50-90}$ $\uparrow$} & \textbf{mIou} $\uparrow$ & \textbf{mAcc} $\uparrow$\\
      
    \midrule
    \multicolumn{8}{l}{\cellcolor{gray!10}\emph{Unlit urban area and rural roads}} \\
    \rowcolor{lowpower} \textbf{LiDAS (Ours)} & 0.6 & 34.6 & 26.9 & 31.0 & 10.8 & \underline{42.8}  & \underline{44.4}  \\        
    \rowcolor{lbpower} Low Beam & 1 & \underline{42.7} & \underline{28.3} & \underline{34.8} & \underline{11.9} & 40.7 & 41.9 \\     
    \rowcolor{lbpower} \textbf{LiDAS (Ours)} & 1 & \textbf{47.1} & \textbf{32.1} & \textbf{40.3} & \textbf{13.1} & \textbf{47.7}  & \textbf{49.4} \\     
    \rowcolor{hbpower} High Beam & 1.8 & 33.9 & 25.8 & 30.0 & 10.4 & 39.6 & 41.1 \\

    \midrule

    \multicolumn{8}{l}{\cellcolor{gray!10}\emph{Urban area with street lights}} \\
    \rowcolor{lowpower} \textbf{LiDAS (Ours)} & 0.6 & 53.0 & 33.4 & 43.8 & 15.6 & 52.7 & 53.3 \\       
    \rowcolor{lbpower} Low Beam & 1 & \textbf{56.0} & \underline{34.3} & \textbf{45.0} & \underline{16.2} & \underline{54.8} & \underline{55.3} \\     
    \rowcolor{lbpower} \textbf{LiDAS (Ours)} & 1 & \underline{54.5} & \textbf{34.6} & \underline{44.2} & \textbf{16.6} & \textbf{55.4} & \textbf{56.0} \\     
    \rowcolor{hbpower} High Beam & 1.8 & 54.0 & 33.9 & 42.6 & 15.1 & 52.2 & 52.9 \\     
    \bottomrule
    \end{tabular}
}
\label{tab:real-world-soma}
\end{table}

Since LiDAS can be bolted onto arbitrary downstream models, we evaluate its compatibility with domain generalization methods. We pair LiDAS with SoMA \cite{yun2025soma} as downstream model and evaluate on our real-world dataset in a zero-shot setting. We report results in \cref{tab:real-world-soma}. SoMA yields significant improvements over YOLOv8L-Worldv2 under low-beam illumination, confirming its DG benefits. Moreover, when combined with our LiDAS policy, we obtain an additional $+5.5$\% mAP and $+7$\% mIoU. These gains underscore LiDAS’s transferability across perception models and the complementarity between active illumination and DG approaches for nighttime safety.

\begin{figure*}

    \centering
    \includegraphics[width=1\textwidth]{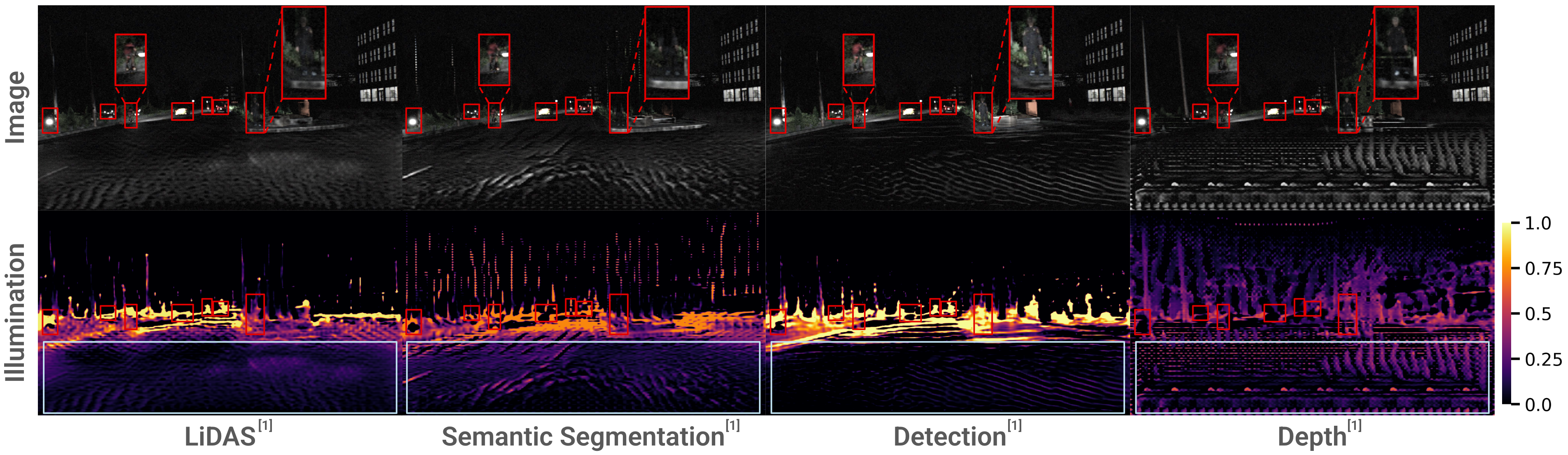}
    \caption{
    \textbf{Task-specific patterns.} The downstream task used at training time directly shapes the learned lighting policy. Detection guide the policy toward sparse, localized highlights over regions of interest, whereas semantic segmentation induces broader, low‑level illumination across the scene. \method leverages both forms of guidance, preserving high contrast on objects while maintaining minimal illumination over the whole scene. In contrast, a depth‑guided policy produces structured light patterns that reveal the scene’s geometry. Method$^{[x]}$ denotes power relative to LB.
    }
    \label{fig:task-specific-pattern}
    \vspace{-8pt}
\end{figure*}

\subsection{Combining downstream tasks}
\label{sec:multitask}

\paragraph{Detection and semantic segmentation.}
As shown in \cref{fig:task-specific-pattern}, the learned illumination is task‑dependent. Training with detectors alone encourages sparse, high‑contrast highlights concentrated on regions of interest. In contrast, training with semantic segmentation induces a low, spatially broad lighting across the scene %
to support correct per‑pixel classification. Combining tasks in \method leverages the complementarity of these signals: the model preserves strong, localized contrasts on salient objects while maintaining sufficient global illumination for scene‑level semantics. This yields a balanced light field that harmonizes patch‑centric and global objectives and benefits both tasks.

\paragraph{Depth estimation.}
\label{sec:depth_task}

As depth is often used in autonomous driving, we train \method with DepthAnything-V2 \cite{depth_anything_v2} as the perception head. \cref{fig:task-specific-pattern} reveals that depth supervision promotes structured, sparsely sampled light, organized in point lines whose size is modulated by distance, yielding an adaptive non‑uniform pattern. This aligns with the structured‑light intuition of LED \cite{demoreau2024led}, yet differs in its adaptive sampling rather than uniform spacing. On our dataset, LED improves depth by $-1.61$ RMSE but requires fine‑tuning the depth network. \method achieves a similar $-1.38$ RMSE gain without any retraining of the downstream head. We report metric details in the suppl. material. However, this depth‑oriented policy conflicts with the area‑based lighting favored by detection and segmentation: depending on the application, we must choose which task maintains performances and which is improved.

\begin{table}[t]
\centering
\caption{\textbf{Ablation study.} Results are reported for \m{LiDAS}{1} with ablated inputs and training strategies, highlighting each component’s contribution to the final performance. }

\resizebox{\linewidth}{!}{
    \begin{tabular}{ccc|c|c|cccc|cc}
    \toprule
     \multicolumn{3}{c|}{\textbf{Inputs}} & \textbf{Residual} & \textbf{Energy} & \multicolumn{4}{c|}{\textbf{Detection}} & \multicolumn{2}{c}{\textbf{Semantic Segmentation}} \\
    
      $I$ & $M_{t-1}$ & $C$ & \textbf{Output} & \textbf{Scheduling} & \textbf{P $\uparrow$} & \textbf{R $\uparrow$} & \textbf{mAP$^{50}$ $\uparrow$} & \textbf{mAP$^{50-90}$ $\uparrow$} & \textbf{mIoU $\uparrow$} & \textbf{mAcc $\uparrow$} \\
      
    \midrule
    \cmark & \xmark & \xmark & \multicolumn{2}{c|}{} & 54.1 & 29.1 & 33.0 & 20.1 & 36.2 & 67.7 \\  
    \cmark & \cmark & \xmark &  \multicolumn{2}{c|}{\raisebox{-3pt}{\smash{\LARGE \cmark}}} & 67.6 & 39.6 & 45.6 & 29.5 & 69.8 & 83.7 \\  
    \cmark & \xmark & \cmark & \multicolumn{2}{c|}{} & 63.5 & 40.2 & 46.0 & 28.8 & 70.5 & 85.3 \\
    \midrule
    & & & \xmark & \cmark & \textbf{68.0} & 38.8 & 46.5 & 29.4 & \textbf{70.7} & \textbf{85.4} \\     
    \multicolumn{3}{c|}{\raisebox{-3pt}{\smash{\LARGE \cmark}}} & \cmark & \xmark & 63.8 & 40.1 & 46.5 & 29.6 & \textbf{70.7} & 85.2 \\     
     &  &  & \cmark & \cmark & 63.9 & \textbf{41.7} & \textbf{46.7} & \textbf{29.7} & \textbf{70.7} & \textbf{85.4} \\     
    \bottomrule
    \end{tabular}
}

\label{tab:ablations}
\end{table}

\subsection{Ablations}
\label{sec:ablations}
\paragraph{Inputs.}
We ablate \method inputs in \cref{tab:ablations}. Adding the previous illumination pattern $M_{t-1}$ improves over RGB‑only by discouraging redundant allocations, providing cues on reflective surfaces, and stabilizing decisions over time. CoordConv $C$ \cite{liu2018intriguing} supplies coarse spatial information, enabling a learned background prior that avoids sky, moderates the near-field to reduce self-glare, and biases coverage toward the horizon. Finally, predicting a residual rather than absolute intensities teaches the model to modify only erroneous regions, simplifying the recursive decision.

\paragraph{Energy scheduling.}
Training directly with a maximal energy budget leads to indiscriminate brightening, especially at high power (\eg, 1.8), where the model illuminates uninformative regions such as the sky because weak energy constraints discourage refinement. Instead, using a linear schedule from $\eta_0$ to $\eta_\mathrm{final}$ over the epochs encourages the model to prioritize essential areas early, followed by later refinement that enhances fine structures and less common regions. Although the performance gap is smaller at lower power, scheduled training consistently leads to better convergence and more targeted illumination.

\section{Conclusion}
\label{sec:conclusion}
We introduce \method, a closed‑loop active‑illumination system that turns high‑definition headlights into a vision actuator, allocating light where it most informs perception. %
Trained only on synthetic data, \method transfers zero‑shot to real‑world closed‑loop driving on unlit roads, where it delivers $+18.7$\% mAP and $+5.0$\% mIoU over low beam. It improves or preserves accuracy while cutting energy use by $40$\%. As a bolt‑on approach, \method transfers across downstream models and complements domain‑generalization methods, offering a cost‑effective path to robust camera‑only perception at night using hardware already present on modern vehicles.

\section*{Acknowledgments}
This work was granted access to the HPC resources of IDRIS under allocation 2025-AD011015334R1 provided by GENCI. This work was partially supported by ELSA - European Lighthouse on Secure and Safe AI funded by the European Union under grant agreement No. 101070617. We would like to thank Amandine Brunetto and the anonymous reviewers for their insightful comments and suggestions. We also thank the Exwayz team for their support on this project, particularly regarding the ground truth of the closed-loop experiment, and the Transpolis team for their help in populating the test tracks.

{
    \small
    \bibliographystyle{ieeenat_fullname}
    \bibliography{main}
}

\clearpage
\setcounter{page}{1}

\setcounter{section}{0}
\setcounter{figure}{0}
\setcounter{table}{0}
\renewcommand\thesection{\Alph{section}}
\renewcommand\thefigure{S\arabic{figure}}    
\renewcommand\thetable{S\arabic{table}}    
\maketitlesupplementary

\begin{table}[b]
\vspace{-8pt}
\centering
\caption{\textbf{Downstream model generalization.} We report YOLO12L \cite{Ultralytics_2023} results, which was not used during training. In this zero‑shot setting, LiDAS improves perception performance, showing its generality. Training downstream models are YOLO11L, YOLOv8L, YOLOv8L-Worldv2, Mask2Former (see \cref{sec:method})}
\resizebox{\linewidth}{!}{
    \begin{tabular}{l|c|cccc}
    \toprule
     \multirow{2}{*}{\textbf{Method}} & \multirow{2}{*}{\textbf{Power}} & \multicolumn{4}{c}{\textbf{Detection}} \\
      &  & \textbf{P $\uparrow$} & \textbf{R $\uparrow$} & \textbf{mAP$^{50}$ $\uparrow$} & \textbf{mAP$^{50-90}$ $\uparrow$} \\
    \midrule
    \multicolumn{1}{l|}{\cellcolor{gray!10}\emph{}} & \multicolumn{1}{l|}{\cellcolor{gray!10}\emph{}}& \multicolumn{4}{l}{\cellcolor{gray!10}\emph{Yolo12L}} \\
    \midrule
    \rowcolor{lowpower} \textbf{LiDAS (Ours)} & 0.6 & \textbf{62.9} & 38.9 & 44.3 & 28.0 \\ 
    \rowcolor{lbpower} Low Beam    & 1 & 53.2 & 31.4 & 34.7 & 19.7 \\
    \rowcolor{lbpower} \textbf{LiDAS (Ours)} & 1 & 60.5 & \underline{40.9} & \textbf{45.0} & \underline{28.4} \\ 
    \rowcolor{hbpower} High Beam   & 1.8 & \underline{62.1} & 36.5 & 42.1 & 25.9 \\
    \rowcolor{hbpower} \textbf{LiDAS (Ours)} & 1.8 & 58.0 & \textbf{41.3} & \underline{44.4} & \textbf{28.7} \\   
    \bottomrule
    \end{tabular}
}
\label{tab:gen_down_sup}
\end{table}

\section{Downstream Tasks Studies}
\paragraph{Combining detection and semantic segmentation.}
As shown in \cref{sec:multitask}, each downstream task induces a distinct illumination policy. We report single‑task results for detection and semantic segmentation in \cref{tab:multitask}. When combined in LiDAS, the two guidance signals regularize the policy and yield further gains for both tasks, highlighting their complementarity. This joint training also produces a more general illumination pattern that remains informative for multiple downstream models.

\begin{table}[t]
\centering
\caption{\textbf{Impact of type of supervision (multi-task, single-task) on performance.} LiDAS trains with both detection and semantic segmentation as downstream tasks, showing improvements for each compared to task‑specific training. Power$\,=\!1$.}
\resizebox{\linewidth}{!}{
    \begin{tabular}{cc|cccc|cc}
    \toprule
     \multicolumn{2}{c|}{\textbf{Training Task}} & \multicolumn{4}{c|}{\textbf{Detection}} & \multicolumn{2}{c}{\textbf{Semantic Segmentation}} \\
      \textbf{Detection} & \textbf{Semantic S.} & \textbf{P $\uparrow$} & \textbf{R $\uparrow$} & \textbf{mAP$^{50}$ $\uparrow$} & \textbf{mAP$^{50-90}$ $\uparrow$} & \textbf{mIoU $\uparrow$} & \textbf{mAcc $\uparrow$} \\
    \midrule
    \checkmark & \xmark & 60.6 & 40.5 & 46.3 & 29.2 & - & - \\
    \xmark & \checkmark & - & - & - & - & 72.0 & 85.1 \\
   \rowcolor{mygray} \checkmark & \checkmark & \textbf{66.9} & \textbf{40.6} & \textbf{47.3} & \textbf{30.0} & \textbf{72.8} & \textbf{87.0} \\
    \bottomrule
    \end{tabular}
}
\label{tab:multitask}
\end{table}

\paragraph{Downstream model generalization.}
Beyond the models in \cref{sec:multitask}, we evaluate LiDAS on additional downstream models unseen during training. In particular, \cref{tab:gen_down_sup} reports results for YOLO12L \cite{Ultralytics_2023}. Consistent with the other models, LiDAS provides informative illumination for YOLO12L and yields consistent performance gains, further demonstrating the generality of the learned policy across multiple downstream models.

\paragraph{Depth estimation.}
We evaluate our method’s ability to enhance depth estimation task in \cref{sec:depth_task}. We compare against LED \cite{demoreau2024led}, another HD‑lighting‑based approach specifically designed to improve depth estimation perception at night. To this end, we reproduce the LED pattern, a checkerboard with 0.25° cells, and apply it using our relighting operator (see \cref{sec:relighting}) on our synthetic dataset.
\Cref{tab:depth} shows that LED reduces RMSE by 1.61\,m and improves the other metrics compared with low‑beam illumination. However, LED requires fine‑tuning the perception models to accommodate its structured illumination pattern. In contrast, LiDAS improves frozen downstream models without any retraining. It learns a more informative illumination pattern for frozen depth model, achieving a comparable 1.38\,m RMSE reduction while using 40\% less power. This demonstrates that our policy can also benefit other tasks when they are included during training. While depth and semantic tasks favor different light structures, LiDAS does not degrade depth performance when optimized for detection/segmentation: baseline accuracy is maintained (RMSE: \m{LiDAS}{0.6}:8.35m, \m{LB}{1}:8.79m,  \m{LiDAS}{1}:8.13m, \m{HB}{1.8}:8.36m). Depending on the application, we choose which task maintains performances and which is improved. 
\begin{table}[t]
\centering
\caption{\textbf{Depth estimation task.} We evaluate LiDAS on depth estimation using DepthAnythingV2 \cite{depth_anything_v2} as the downstream model. We compare against LED \cite{demoreau2024led}, which requires fine‑tuning the perception model, whereas our results use the pretrained DepthAnythingV2 in a zero‑shot setting.}

\resizebox{\linewidth}{!}{
    \begin{tabular}{l|c|cccc}
    \toprule
     \textbf{Method} & \textbf{Power} & \textbf{RMSE $\downarrow$} & \textbf{Abs Rel $\downarrow$} & \textbf{SiLog $\downarrow$} & $\mathbf{\delta_1} $ $\uparrow$\\
    \midrule
    \multicolumn{6}{l}{\cellcolor{gray!10}\emph{With fine-tuning}} \\
    \rowcolor{lbpower} Low Beam    & 1 & 5.14 & 0.117 & 0.157 & 0.903\\ 
    \rowcolor{lbpower} LED \cite{demoreau2024led} & 1 & \textbf{3.53} & \textbf{0.066} & \textbf{0.103} & \textbf{0.950} \\
    \rowcolor{hbpower} High Beam   & 1.8 & 4.79 & 0.107 & 0.148 & 0.909 \\

    \midrule
    \multicolumn{6}{l}{\cellcolor{gray!10}\emph{Zero-shot}} \\
    \rowcolor{lowpower} \textbf{LiDAS (Ours)} & 0.6  & \underline{7.41} & \underline{0.178} & \underline{0.225} & \underline{0.745} \\ 
    \rowcolor{lbpower} Low Beam    & 1 & 8.79 & 0.305 & 0.309 & 0.343\\ 
    \rowcolor{lbpower} LED \cite{demoreau2024led} & 1 & 11.3 & 0.250 & 0.381 & 0.586 \\ 
    \rowcolor{lbpower} \textbf{LiDAS (Ours)} & 1  & 7.68 & \underline{0.178} & 0.233 & 0.732 \\ 
    \rowcolor{hbpower} High Beam   & 1.8 & 8.36 & 0.297 & 0.294 & 0.358 \\ 
    \rowcolor{hbpower} \textbf{LiDAS (Ours)} & 1.8 & \textbf{7.23} & \textbf{0.165} & \textbf{0.216} & \textbf{0.777} \\

    \bottomrule

    \end{tabular}
}
\vspace{-8pt}

\label{tab:depth}
\end{table}

\vspace{-10pt}

\section{Performance Over Distance}
\label{sec:distance}
\begin{figure}[t]

    \centering
    \includegraphics[width=0.8\columnwidth]{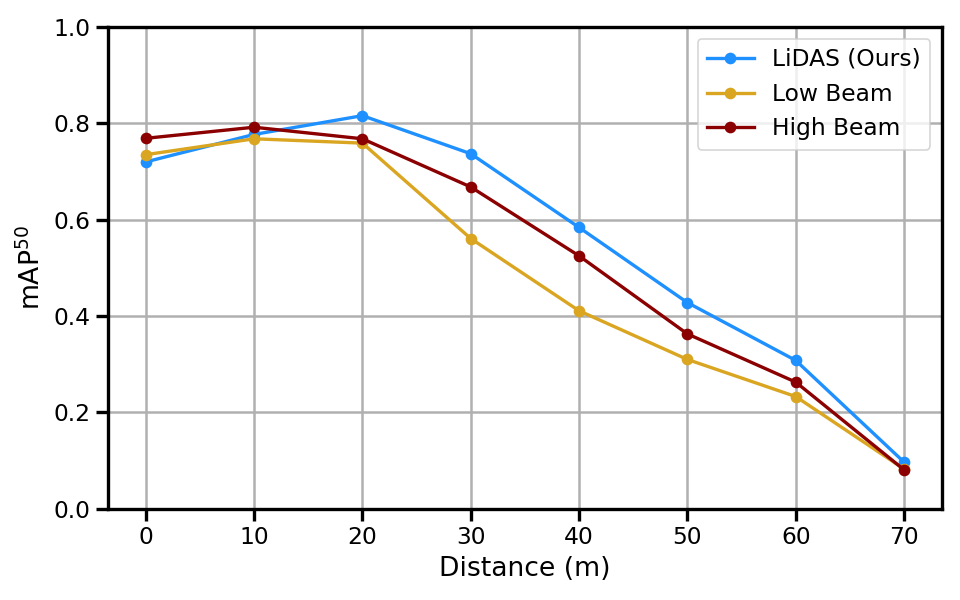}
    \caption{
    \textbf{Performance over distance.} LiDAS improves detection in the 20-60\,m range, a safety‑critical band for applications such as autonomous emergency braking.
    }
    \label{fig:perf-vs-dist}
\end{figure}

\cref{fig:perf-vs-dist} shows that \m{LiDAS}{1} leads in the 20-60\,m band, a safety‑critical region for early Autonomous Emergency Braking (AEB) triggers. Methods are statistically similar at 0-20\,m, where all are sufficiently bright and the main concern is self‑glare. Beyond 70\,m, performance converges again due to limited energy on small, distant targets. Contrary to intuition, HB does not dominate at long range: its light pattern does not illuminate lateral areas and thus primarily benefits centered targets. In contrast, LiDAS widens illumination around the horizon, improving accuracy for all far objects.

\section{Performance Over Time}
\begin{figure}[b]

    \centering
    \includegraphics[width=1\columnwidth]{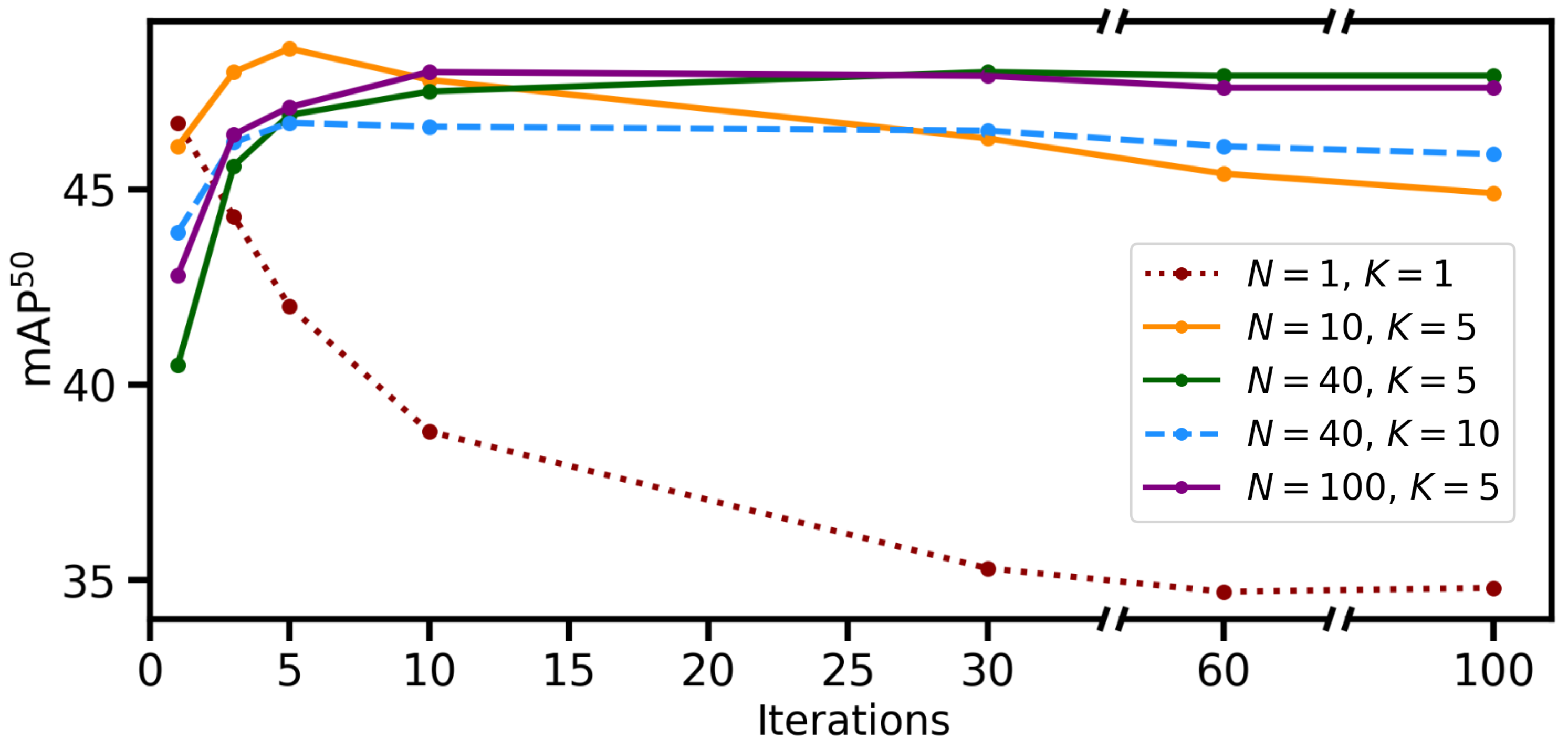}
    \caption{
    \textbf{Performance over time.} Increasing the step count $N$ mitigates performance drift. Adjusting $K$ balances exposure to the initial random pattern and the model’s own light during training.
    }
    \label{fig:autoregressive-study}
\end{figure}

We study the sequential refinement parameters (see \cref{sec:autoregressive}) in \cref{fig:autoregressive-study}. 
With $K{=}1$, the model sees only the initial random pattern $M_0$ during training and thus degrades quickly once exposed to its own illumination at test time. Increasing the unroll length $N$ mitigates this drift: $N{=}10$ maintains performance longer, $N{=}40$ effectively stabilizes it, while $N{=}100$ brings no further benefit, yet substantially increases training time. 
Thus, making $N{=}40$ the best compute-accuracy trade‑off. Varying $K$ also matters as it sets the balance between the random start $M_0$ and LiDAS-generated light seen during training. We find $K{=}5$  best balances robustness to arbitrary initial light fields with stability under self‑generated illumination. We note that LiDAS requires a brief warm‑up to reach peak performance: 20-30 iterations suffice, corresponding to 1-2\,s of initialization at vehicle startup, which is negligible in our active perception setting.

\section{Evaluation on Public Datasets}
\begin{figure}[t]

    \centering
    \includegraphics[width=0.8\columnwidth]{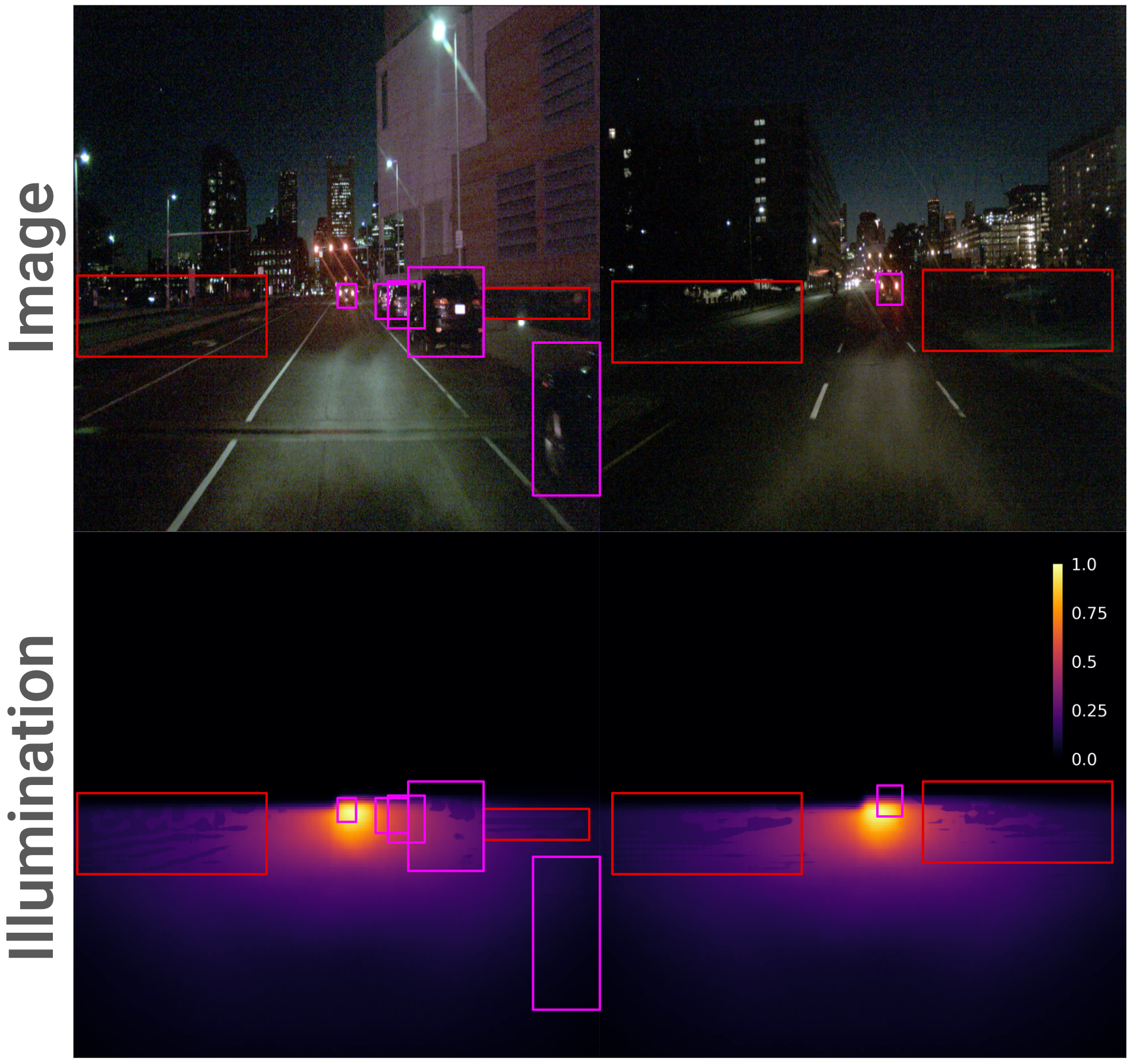}
    \caption{
    \textbf{Qualitative results on nuImages (night).} Red boxes indicate darkened areas. Purple boxes denote ground‑truth objects. LiDAS selectively reduces illumination only in uninformative regions of the scene.
    }
    \label{fig:nuImages}
    \vspace{-8pt}
\end{figure}

Evaluating on existing datasets is challenging because we cannot physically alter the car’s illumination as captured by the camera. 
We assume that night frames are LB-illuminated ($M_{\text{LB}}$), which is the expected setup in urban areas. Since captured images cannot gain photons, we only attenuate regions where our model requests less light than the LB. Accordingly, we apply our relighting operator to simulate new lighting under this darken-only constraint: Given an input $I_{\text{LB}}$ and a desired illumination map $M\in[0,1]$, we interpolate with a black image:

\begin{equation}
M' = 1 - \max(M_{\text{LB}} - M,\, 0), \quad 
\hat{I}' = I_{\text{LB}} \odot M'.
\end{equation}
This protocol cannot reveal new details, it can only remove information. \Cref{tab:nuimages} shows that even under this unfavorable setting, \mm{LiDAS} performances are comparable to the \mm{Low beam} on the nuImages \cite{caesar2020nuscenes} dataset while reducing power by 30\%. It demonstrates that it has learned to unlit only the uninformative regions of the scene as shown in \cref{fig:nuImages}. Following the same protocol, we report results on BDD100K \cite{yu2020bdd100k} in \cref{tab:bdd100k} and NightCity \cite{tan2021night} in \cref{tab:nightcity}. Results are consistent with performances observed on nuImages, reducing power consumption by 20\% while maintaining performances.  In an active-illumination system, the energy saved from dimmed areas would be reallocated to informative area of the scene, which could further improve perception. 

\section{Public Baselines}
As a bolt-on system, LiDAS is compatible with arbitrary perception architectures, we integrated the DTP \cite{wei2023disentangle} SOTA nighttime model as a downstream head on our closed-loop dateset. Results are reported in \cref{tab:dtp}. LB and HB represent DTP’s native performance under standard lighting. LiDAS consistently improves metrics for this nighttime-specialized model in a zero-shot setting, demonstrating it's compatibility with both DA/DG methods and models dedicated for nighttime perception.

\begin{table}[b]
\vspace{-8pt}
\centering
\caption{\textbf{Results using DTP in closed-loop.} LiDAS improves performances of the DTP nighttime-specialized model over the standard illumination. }

\resizebox{0.8\linewidth}{!}{
    \begin{tabular}{l|c|cc}
    \toprule
     \multirow{2}{*}{\textbf{Method}} & \multirow{2}{*}{\textbf{Power}} & \multicolumn{2}{c}{\textbf{Semantic Segmentation}} \\
        
      &  & \textbf{mIoU $\uparrow$} & \textbf{mAcc $\uparrow$} \\
      
    \midrule
    \rowcolor{lowpower} \textbf{LiDAS (Ours)} & 0.6 & \underline{19.0} & \underline{19.3} \\   
    \rowcolor{lbpower} Low Beam & 1 & 18.3 & 18.5 \\
    \rowcolor{lbpower} LiDAS (Ours) & 1 & \textbf{23.2} & \textbf{23.9} \\
    \rowcolor{hbpower} High Beam & 1 & 18.0 & 18.3 \\
    \bottomrule
    \end{tabular}
}

\label{tab:dtp}
\end{table}

\section{Runtime Details}
During training we used an RTX 4090 on which the inference takes 6.8\,ms. For our real-world deployment, we used a laptop-GPU (RTX 2000 Ada) and achieved 32\,FPS (31.2\,ms) at full and 97\,FPS (10.3\,ms) at half resolution. Our parallel pipeline, handling camera I/O, LiDAS inference, and headlight projection ($\le$15\,ms), maintains a total system latency of $\le$46.2\,ms, which has proven to be responsive enough. LiDAS runs in parallel with the perception stack, adding no latency to downstream models. While the size of the model was not our focus, standard compression techniques (distillation, pruning) can optimize it for ECUs. Successful real-world tests shows the system's feasibility.

\section{Relighting Operator Limitations}
The linear interpolation is a differentiable proxy designed for computational efficiency only during training. We acknowledge that it does not model complex physical effects (specular reflections, camera ISP non linearities...), still, its effectiveness is validated by LiDAS improvement on real-world perception in zero-shot transfer from synthetic data. It shows the operator provides a sufficient signal to learn a robust illumination policy.

\section{Lighting Regulations}
Our policy does not explicitly prevent glare toward other road users, so practical deployment will likely require integration with anti‑glare systems \cite{fleet_programmable_2014}. 
LiDAS is compatible with these methods, which could mask sensitive regions in our output. We consider this a critical safety  trade-off: with $77$\% of pedestrian crashes occurring at night\footnote{NHTSA. Traffic Safety Facts: 2023 Data-Pedestrians. US Dept. of Transp. 2025. Report No. DOT HS 813 727.}, LiDAS prioritizes illumination for life-saving detection, in less-populated very low-light environments. 
Regulations remain a near‑term barrier: many jurisdictions do not yet authorize fully dynamic HD headlight functions, and LiDAS may need adaptations (e.g., exclusion zones, intensity and update‑rate limits) as rules evolve. Still, regulation is evolving toward dynamic HD lighting: active working groups are studying the use of light to assist ADAS, it will help define the balance between visibility and glare. 
\newpage

\begin{table}[b]
\vspace{-8pt}
\centering
\caption{\textbf{Results on nuImages (night).} LiDAS achieves performance comparable to Low Beam while using $30$\% less energy, indicating that it suppresses light only in uninformative regions.}

\resizebox{\linewidth}{!}{
    \begin{tabular}{l|c|cccc}
    \toprule
     \multirow{2}{*}{\textbf{Method}} & \multirow{2}{*}{\textbf{Power}} & \multicolumn{4}{c}{\textbf{Detection}} \\
        
      &  & \textbf{P $\uparrow$} & \textbf{R $\uparrow$} & \textbf{mAP$^{50}$ $\uparrow$} & \textbf{mAP$^{50-90}$ $\uparrow$}\\
      
    \midrule
    \rowcolor{lowpower} \textbf{LiDAS (Ours)} & 0.7 & 61.1 & 23.7 & 41.0 & 23.8  \\   
    \rowcolor{lbpower} Low Beam & 1 & 62.1 & 24.8 & 43.0 & 26.3   \\
    \bottomrule
    \end{tabular}
}

\label{tab:nuimages}
\end{table}

\begin{table}[b]
\vspace{-8pt}
\centering
\caption{\textbf{Results on BDD100K (night).} LiDAS achieves performance comparable to Low Beam while using $20$\% less energy, indicating that it suppresses light only in uninformative regions.}

\resizebox{\linewidth}{!}{
    \begin{tabular}{l|c|cccc}
    \toprule
     \multirow{2}{*}{\textbf{Method}} & \multirow{2}{*}{\textbf{Power}} & \multicolumn{4}{c}{\textbf{Detection}} \\
        
      &  & \textbf{P $\uparrow$} & \textbf{R $\uparrow$} & \textbf{mAP$^{50}$ $\uparrow$} & \textbf{mAP$^{50-90}$ $\uparrow$}\\
      
    \midrule
    \rowcolor{lowpower} \textbf{LiDAS (Ours)} & 0.8 & 51.1 & 23.2 & 36.1 & 24.0  \\   
    \rowcolor{lbpower} Low Beam & 1 & 50.0 & 24.2 & 36.1 & 24.6   \\
    \bottomrule
    \end{tabular}
}

\label{tab:bdd100k}
\end{table}

\begin{table}[b]
\vspace{-8pt}
\centering
\caption{\textbf{Results on NightCity.} LiDAS achieves performance comparable to Low Beam while using $20$\% less energy, indicating that it suppresses light only in uninformative regions. We used SoMA as downstream model for this experiment, demonstrating its compatibility.}

\resizebox{0.8\linewidth}{!}{
    \begin{tabular}{l|c|cc}
    \toprule
     \multirow{2}{*}{\textbf{Method}} & \multirow{2}{*}{\textbf{Power}} & \multicolumn{2}{c}{\textbf{Semantic Segmentation}} \\
        
      &  & \textbf{mIoU $\uparrow$} & \textbf{mAcc $\uparrow$} \\
      
    \midrule
    \rowcolor{lowpower} \textbf{LiDAS (Ours)} & 0.8 & 51.4 & 74.5 \\   
    \rowcolor{lbpower} Low Beam & 1 & 51.5 & 74.7 \\
    \bottomrule
    \end{tabular}
}

\label{tab:nightcity}
\end{table}

\end{document}